\documentclass[runningheads]{llncs}
\usepackage[T1]{fontenc}
\usepackage{graphicx}
\usepackage{booktabs}

\usepackage{amsmath,amsfonts,amssymb,amsthm,dsfont}

\usepackage{hyperref}
\usepackage{url}
\usepackage{todonotes}
\usepackage{float}
\usepackage{url}

\usepackage{multirow}
\usepackage{multicol}
\usepackage{booktabs} 
\usepackage{colortbl}
\def\redc{\cellcolor[HTML]{FF999A}}
\def\orangec{\cellcolor[HTML]{FFCC99}}
\def\yellowc{\cellcolor[HTML]{FFF8AD}}
\def\R{\mathbb{R}}

\def\r{{\bf r}}

\def\G{\mathcal{G}}
\def\N{\mathcal{N}}
\def\m{\mathrm{m}}
\def\x{ {\bf x} }

\def\E{\mathcal{E}}

\def\our{REdiSplats}

\usepackage{wrapfig}

\theoremstyle{plain}

\usepackage{pifont}
\usepackage{mwe}

\usepackage{hyperref}

\usepackage{color}

\usepackage[capitalize,noabbrev]{cleveref}

\usepackage{mathtools}
\usepackage{tabularx}
\usepackage{algpseudocode}
\usepackage{algorithm}
\algnewcommand\algorithmicforeach{\textbf{for each}}
\algdef{S}[FOR]{ForEach}[1]{\algorithmicforeach\ #1\ \algorithmicdo}

\makeatletter
\newcommand{\multiline}[1]{%
  \begin{tabularx}{\dimexpr\linewidth-\ALG@thistlm}[t]{@{}X@{}}
    #1
  \end{tabularx}
}
\makeatother

\begin{document}

\title{\our{}: Ray Tracing for Editable Gaussian Splatting}

\titlerunning{\our{}: Ray Tracing for Editable Gaussian Splatting}

\author{
Krzysztof Byrski\inst{1}
\and
Grzegorz Wilczyński\inst{1}
\and
Weronika Smolak-Dyżewska\inst{1}
\and\\
Piotr Borycki\inst{1}
\and
Dawid Baran\inst{1}
\and
Sławomir Tadeja\inst{2}
\and\\
Przemysław Spurek\inst{1}
}

\institute{
Faculty of Mathematics and Computer Science, Jagiellonian University, Poland\\\email{przemyslaw.spurek@uj.edu.pl}
\and
Department of Engineering, University of Cambridge, Cambridge, United Kingdom\\
}

\maketitle              

\begin{abstract}
Gaussian Splatting (GS) has become one of the most important neural rendering algorithms. GS represents 3D scenes using Gaussian components with trainable color and opacity. This representation achieves high-quality renderings with fast inference. 
Regrettably, it is challenging to integrate such a solution with varying light conditions, including shadows and light reflections, manual adjustments, and a physical engine. Recently, a few approaches have appeared that incorporate ray-tracing or mesh primitives into GS to address some of these caveats. However, no such solution can simultaneously solve all the existing limitations of the classical GS. Consequently, we introduce \our{}, which employs ray tracing and a mesh-based representation of flat 3D Gaussians. In practice, we model the scene using flat Gaussian distributions parameterized by the mesh. We can leverage fast ray tracing and control Gaussian modification by adjusting the mesh vertices. Moreover, \our{} allows modeling of light conditions, manual adjustments, and physical simulation. Furthermore, we can render our models using 3D tools such as Blender or Nvdiffrast, which opens the possibility of integrating them with all existing 3D graphics techniques dedicated to mesh representations. 
The code is available at \url{https://github.com/KByrski/REdiSplats}.
\end{abstract}

\section{Introduction}

Gaussian Splatting (GS) \cite{kerbl20233d} allows the generation of novel views of 3D objects using 2D images. GS represents a 3D scene by Gaussian distributions with colors, and opacity. Such a representation provides high-quality renders and fast rendering time. These properties are obtained by using restoration instead of ray tracing. In practice, Gaussian components are projected onto 2D planes, which is highly efficient. Unfortunately, such a solution has certain limitations. GS-based objects require specialized rendering engines that do not respond to varying light conditions or mesh-based objects. These limitations stem from the employing restoration rather than ray tracing \cite{glassner1989introduction}.

Recognizing the inherent limitations of standard GS, recent works have explored hybrid approaches that combine it with ray tracing. These efforts aim to leverage the strengths of both techniques. One such method, 3D Gaussian Ray Tracing (3DGRT) \cite{moenne20243d}, introduces bounding primitives surrounding each Gaussian to enable the application of ray tracing to these simplified geometries rather than the Gaussians themselves. Another approach, RaySplats~\cite{byrski2025raysplats}, proposes using elliptical approximations of Gaussians to facilitate ray-surface intersection calculations. Such models achieve a reconstruction comparable to that of GS, allowing integration with lighting conditions and meshes. Instead of adding ray tracing to GS, we can use a direct mesh representation for 3D object rendering. In MeshSplats~\cite{tobiasz2025meshsplats}, the authors propose a transformation from GS to mesh that enables the rendering of GS objects in tools like Blender and Nvdiffrast. LinPrim \cite{von2025linprim} utilizes linear primitives such as octahedra and tetrahedral to perform differentiable volumetric rendering. This method favors mesh-based primitives rather than Gaussians. 

\begin{figure}[!t]
    \centering
        \begin{center}
            \our{} for light reflections and mesh interaction\\
        \includegraphics[width=0.99\linewidth, trim=0 0 0 0, clip]{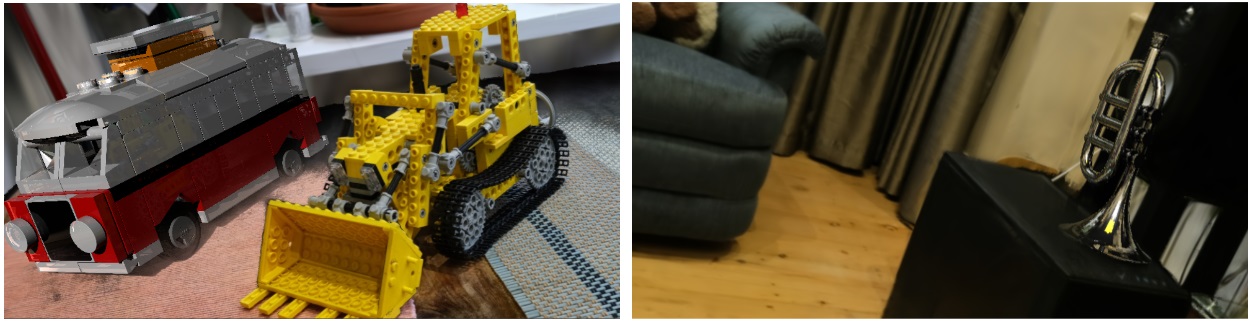}
        \\
        \our{} combined with physical engine\\[1mm]
         \includegraphics[width=0.32\linewidth]{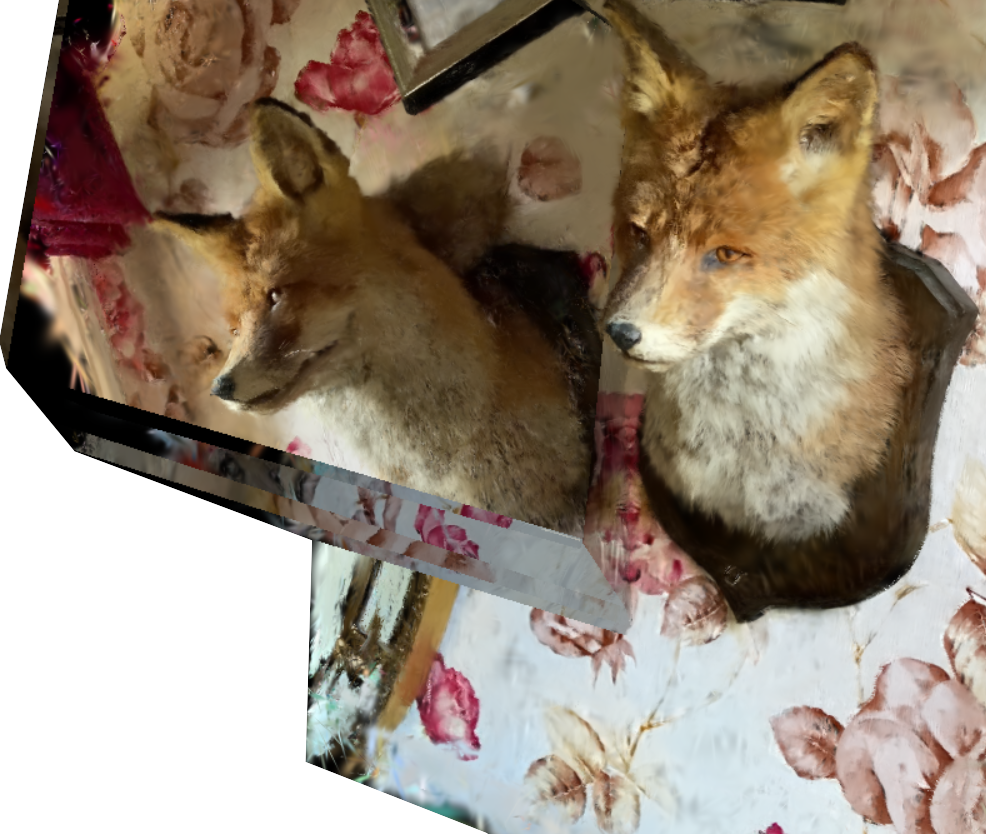}
         \includegraphics[width=0.32\linewidth]{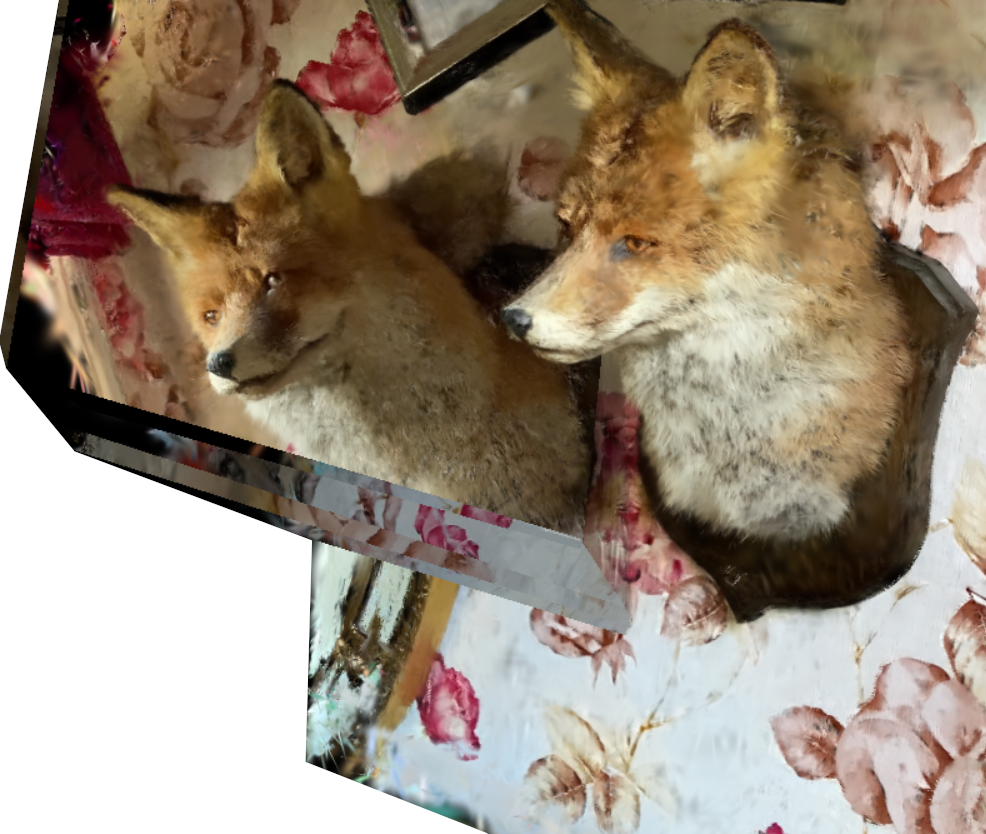}
         \includegraphics[width=0.32\linewidth]{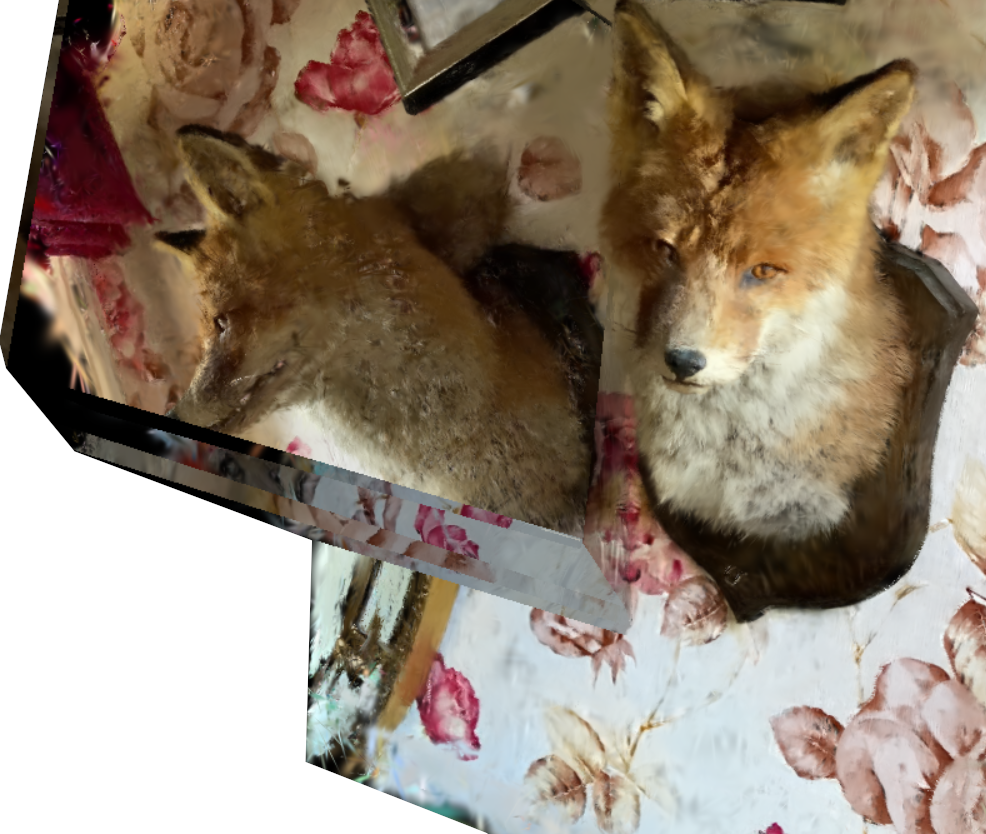}

        \end{center} 
    \caption{\our{} model allows to model varying light conditions, including shadows, light reflections and mirror reflection. Moreover, we can edit the scene manually or using a physical ending.}
    \label{fig:tesser}
\end{figure}

The above solutions address selected constraints of classical vanilla GS and belong to two main groups. First, ray-tracing-based approaches incorporate light conditions and meshes. In contrast, the mesh-based models in the second group enable manual modification and utilize physical engines. However, none of these solutions provides the benefits bestowed by all of these properties. 

\begin{figure}[!t]
    \centering
        \begin{center}
            \our{} in Blender simulations\\
        \includegraphics[width=0.99\linewidth]{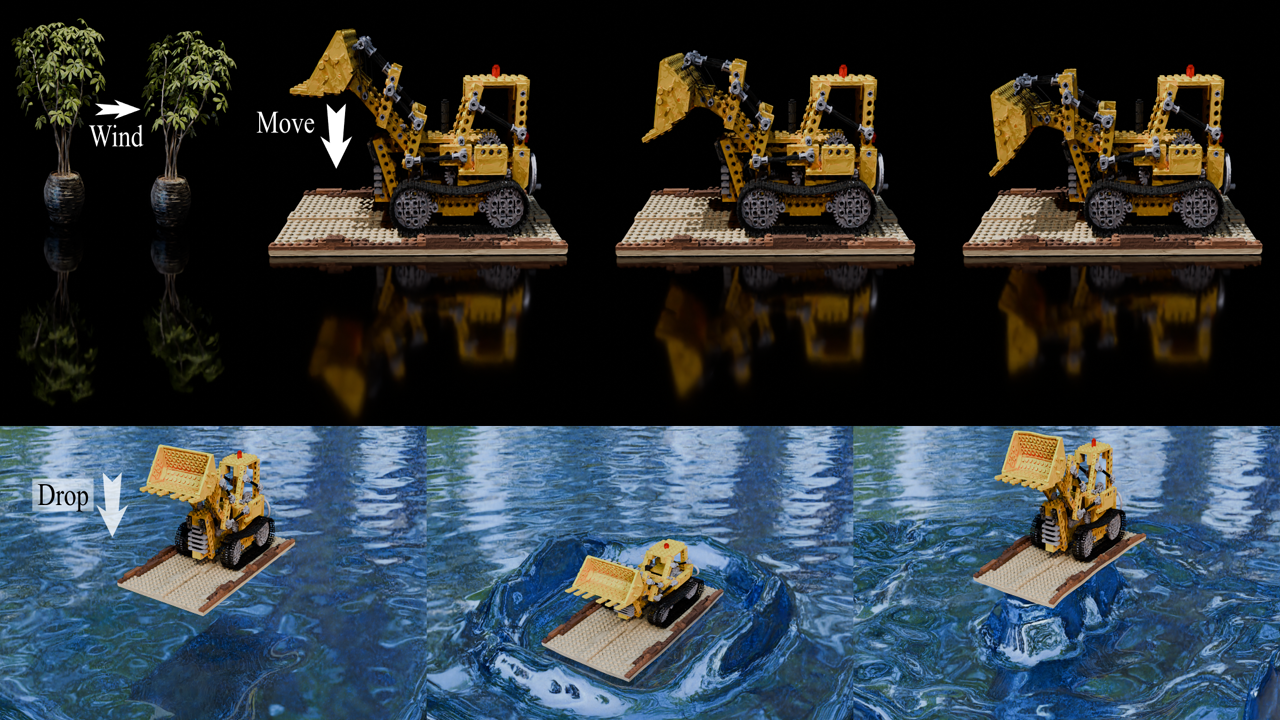}
        \end{center} 
    \caption{\our{} incorporates ray tracing into the 3D~Gaussian Splatting framework. The figure presents a physical simulation model rendered in Blender.}
    \label{fig:blender}
\end{figure}

Consequently, we propose a new model \our{}, which uses flat Gaussians and its mesh approximation. Flat Gaussian distributions can be effectively utilized in GS by adjusting the Gaussian scaling component to have a zero value on one axis \cite{waczynska2024games}. Thus, a flat Gaussian can be understood as 2D ellipse, easily approximated with a mesh. Such representation has two main properties. First, we can use it for ray tracing capabilities. By combining the GS engine with NVIDIA OptiX \cite{parker2010optix}, we obtain a model capable of simulating shadows, light, and mirror reflections. Secondly, the mesh face representation can be used for Gaussian modification \cite{waczynska2024games,waczynska2025d,borycki2024gasp,waczynska2024mirage}. The mesh vertices can be modified manually or by a physical engine to achieve the position of the Gaussian component, as shown in Fig.~\ref{fig:tesser}. Moreover, we can transform a flat Gaussian into a mesh-based representation, which can be rendered using existing rendering tools like Blender and Nvdiffrast as presented in Fig.~\ref{fig:blender}. 

In summary, we introduce \our{}, which is an innovative differential rendering method for 3D GS that incorporates ray tracing in both the training and inference stages. Our model is capable of the following:
\begin{itemize}
    \item \our{} enables handling lighting effects (such as reflections, shadows, transparency) and allows combining GS models with mesh-based models;
    \item \our{} allows manual edition and physical simulation of  GS;
    \item \our{} can render in existing rendering tools like Blender and Nvdiffrast.
\end{itemize}

\section{Related Works}

We structure our review into two segments. Initially, we provide an overview of the novel-view synthesis, followed by a detailed examination of the ray-tracing techniques.

GS allows modeling a 3D scene using a set of Gaussian distributions \cite{kerbl20233d}. Thanks to an efficient projection algorithm, it can model novel, high-resolution views. Unfortunately, rasterization does not allow modeling light conditions such as reflections, shadows, or transparency. It is also not obvious how to integrate mesh objects with GS.

\begin{wrapfigure}{r}{0.5\textwidth}
\vspace{-0.5cm}
  \begin{center}
    \centering
    \includegraphics[width=0.95\linewidth]{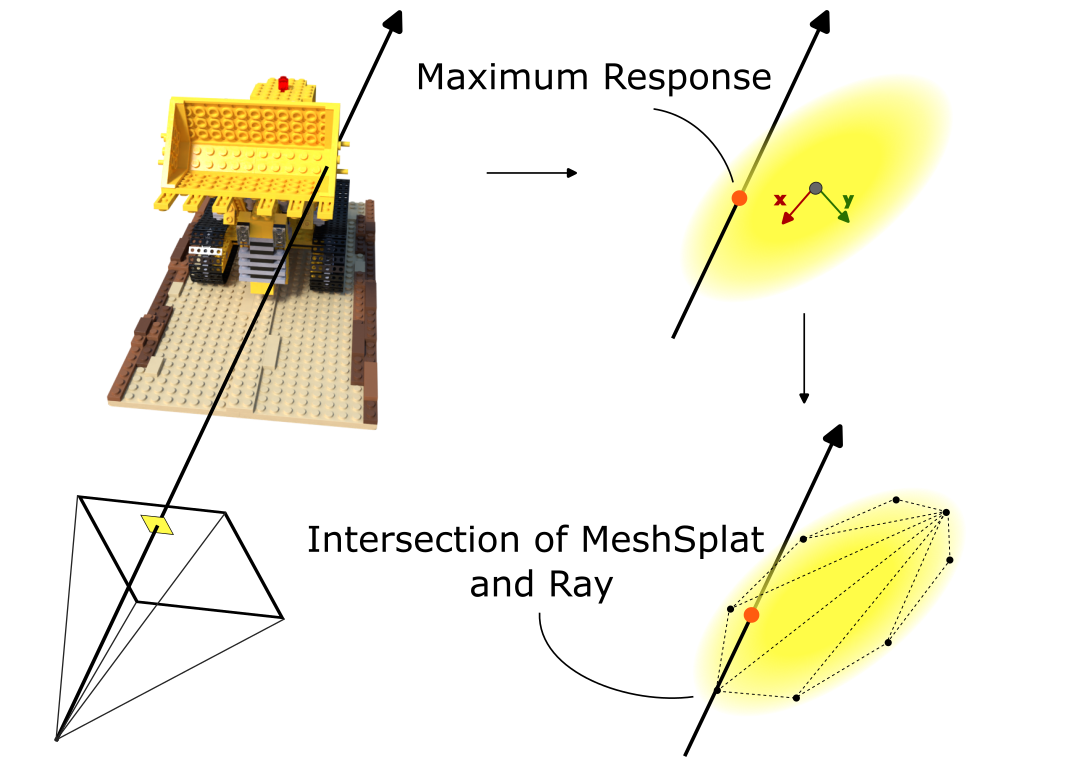}
    \caption{\our{} (our) uses ray-tracing-based approach. We need two points with regard to rays passing through Gaussian distributions. We use flat GS to utilize mesh to approximate such primitives, with the mesh-ray intersection being very efficient.}
    \label{fig:ray_mesh}
\end{center}
\vspace{-0.9cm}
\end{wrapfigure}

The challenge of incorporating efficient and accurate ray tracing with 3D Gaussian primitives has been tackled by introducing 3D Gaussian Ray Tracing (3DGRT) \cite{moenne20243d}. This technique utilizes a differentiable ray tracer, enabling direct ray-Gaussian intersection calculations and subsequent radiance evaluation. Notably, the implementation leverages the NVIDIA OptiX framework \cite{parker2010optix}, resulting in enhanced rendering quality compared to traditional rasterization methods. Alternatively, RaySplats~\cite{byrski2025raysplats} approximate Gaussians using ellipses instead of bounding primitives, enabling GS-based objects to integrate with lighting conditions and meshes while maintaining reconstruction quality comparable to that of GS. However, the above models are restricted to Gaussian primitives, limiting compatibility with more generalized distributions \cite{condor2025don,hamdi2024ges,kasymov2024neggs}.

Although Gaussian-based rendering has seen significant progress, limitations remain in its ability to handle complex lighting effects and 3D scene reconstruction. Inter-Reflective Gaussian Splatting (IRGS) \cite{gu2024irgs} represents an attempt to address inter-reflection modeling by extending the 2D Gaussian Splatting (2DGS) framework \cite{huang20242d} with a differentiable ray-tracing approach. This method effectively combines the rendering equation with 2D flat Gaussians to enable relighting. Nevertheless, the fundamental reliance on 2D primitives within IRGS restricts its applicability for tasks that require complete 3D scene representation and integration with traditional mesh-based models.

Accurately representing complex reflections in real-world scenes remains a significant challenge for GS. In response, EnvGS \cite{xie2024envgs} proposes a novel approach that integrates ray tracing with environment Gaussian primitives. This integration enables the capture of near-field and high-frequency reflections, overcoming the inherent limitations of environment maps, which rely on distant lighting approximations. The method's strength lies in combining these specialized Gaussian primitives with the base 3D GS representation, achieving detailed and real-time rendering. However, this is achieved through a two-stage optimization process, with ray tracing applied post-3DGS pre-training.

Rather than incorporating ray tracing into GS, an alternative approach is to use a direct mesh representation for 3D object rendering. MeshSplats \cite{tobiasz2025meshsplats} transforms GS to meshes, allowing GS objects to be rendered within tools like Blender and Nvdiffrast. Similarly, LinPrim \cite{von2025linprim} uses linear primitives such as octahedra and tetrahedral for differentiable volumetric rendering, prioritizing mesh-based primitives over Gaussian representations.

The ray-tracing models solve problems with light reflections and mesh integration. Unfortunately, it is not obvious how to integrate such models with physical engines or allow their manual modifications. \our{} solve such problems by using flat Gaussians in ray tracing GS. Such parametrization allows us to easily modify GS-based objects.  

\begin{figure}[!t]
    \centering
        \begin{center}
            \our{} with mirror and light reflections\\
        \includegraphics[width=0.99\linewidth, trim=0 0 0 0, clip]{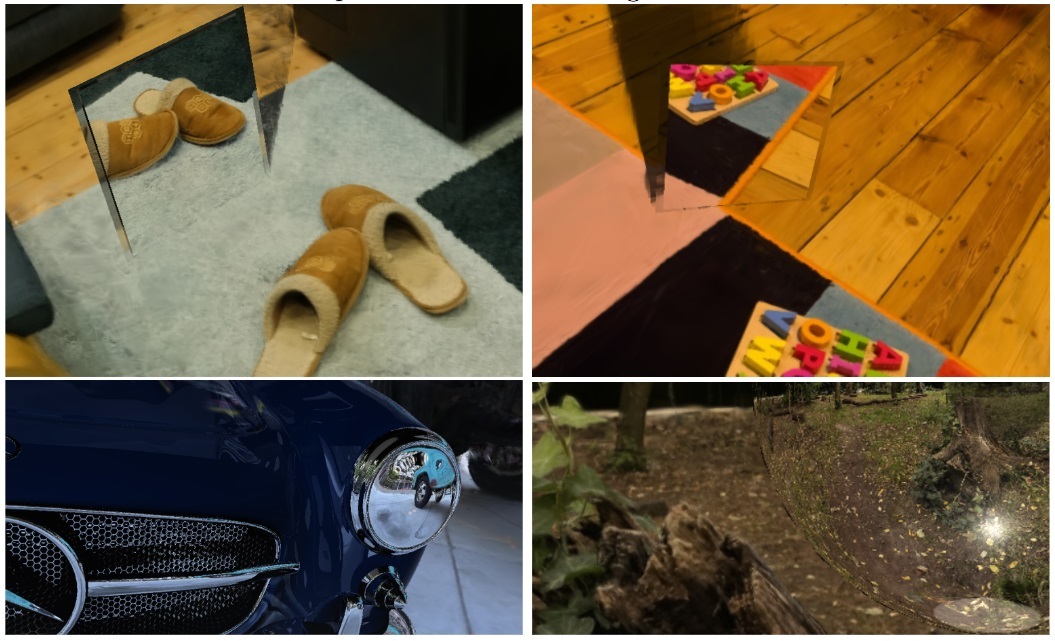}    
        \end{center} 
    \caption{\our{} allows to model mirror reflection and light conditioning thanks to ray tracing model.}
    \label{fig:gs_mirror}
\end{figure}

\section{\our{}--Ray Tracing for Editable GS}

Here, we present \our{}, which allows editing ray tracing-based GS. We begin with the classical GS and its flat version. We then introduce the mesh approximation of a flat Gaussian. The idea is inspired by MeshSplats \cite{tobiasz2025meshsplats}, but we use fewer faces, which is essential for the efficient implementation of ray tracing. Next, we present the process of determining the intersection of a ray with a flat Gaussian represented by a mesh. It is crucial to note that identifying Gaussians that intersect with a given ray is essential for aggregating colors along that ray, as shown in Fig.~\ref{fig:ray_mesh}. In 3DGRT \cite{moenne20243d} and RaySplats~\cite{byrski2025raysplats}, we need two points: one for identifying the Gaussian and the second for opacity modulation. In \our{}, one point fulfills both roles. 

\subsection{3D Gaussian Splatting} 
Gaussian Splatting (GS) \cite{kerbl20233d} models 3D scenes with a collection of Gaussians:
\begin{equation}\label{eq:3dgs}
    \mathcal{G} = \{ (\mathcal{N}(\mathbf{m}_i,\Sigma_i), \hat{\alpha}_i, c_i) \}_{i=1}^{n},
\end{equation}
where trainable parameters $\mathbf{m}_i$, $\Sigma_i$, $\hat{\alpha}_i$, and $c_i$ are the covariance, position (mean), opacity, and color of the $i$-th component. The color representation utilizes the Spherical Harmonics (SH) \cite{fridovich2022plenoxels,muller2022instant}.  
GS employs the covariance matrix factorization:
\begin{equation}\label{eq:factorization}
    \Sigma = R S S^T R^T,
\end{equation}
where $R$ represents the rotation matrix, and $S$ is a diagonal matrix containing the scaling parameters.  

\begin{figure}[!t]
    \centering
        \begin{center}

        \our{} combine with mesh-based objects\\
        \includegraphics[width=0.99\linewidth, trim=0 0 0 0, clip]{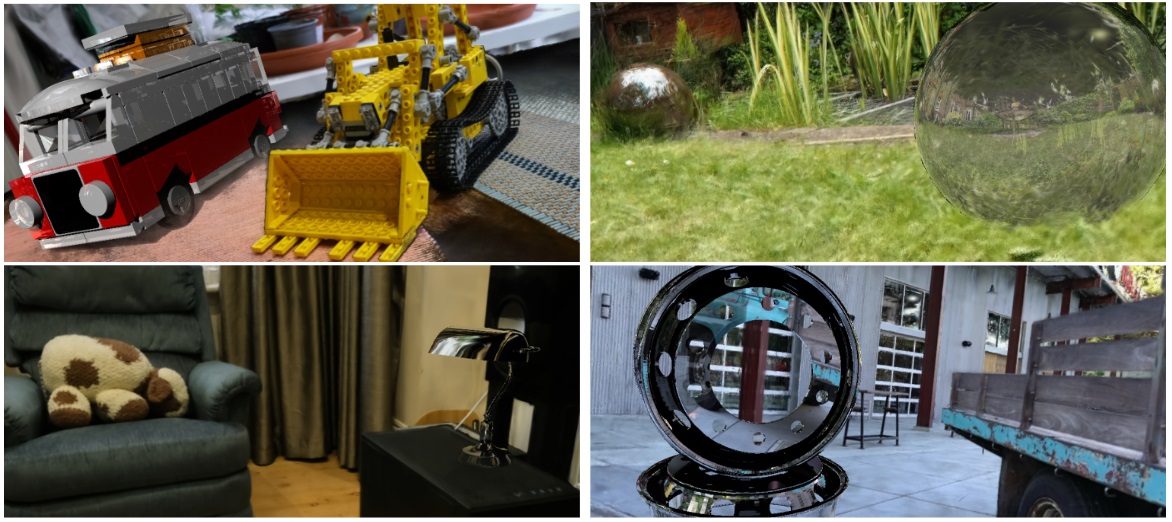}
        \end{center} 
    \caption{\our{} use ray tracing ending instead of restoration. Therefore, we can build scenes from GS and add a mesh-based element that interacts with the scene.}
    \label{fig:gs_mesh}
\end{figure}

The GS algorithm maps Gaussian distributions onto the image plane by blending colors for individual pixels. This is achieved by sampling from the respective Gaussian distributions $\mathcal{N}(\mathbf{m}_i, \Sigma_i)$, where overlapping points contribute to the final color of the pixels. This approach follows the method described in \cite{kopanas2021point,kopanas2022neural}. The pixel color is computed as follows \cite{yifan2019differentiable}:
\begin{equation}\label{equ:color}
    C = \sum_{i=1}^N c_i \alpha_i \prod_{j=1}^{i-1} (1-\alpha_j),
\end{equation}
where $\alpha_i$ is determined by evaluating a 2D Gaussian with covariance matrix $\Sigma$, multiplied by a learned per-Gaussian opacity $\hat{\alpha}_i$.

This formula results from point-based alpha blending~\cite{kerbl20233d}, where the color of a ray is defined as:
\begin{equation}\label{eq:color1}
    C = \sum_{i=1}^N T_i (1-\exp(-\sigma_i\delta_i))c_i,
\end{equation}
with

\begin{equation}\label{eq:color2}
    T_i = \exp\left(-\sum_{j=1}^{i-1} \sigma_j \delta_j\right),
\end{equation}
where $\sigma_i$ represents the density, $T_i$ the transmittance, and $c_i$ the color of samples collected along the ray at intervals $\delta_i$.

\begin{wrapfigure}{r}{0.5\textwidth}
\vspace{-1.5cm}
  \begin{center}
    \includegraphics[width=0.8\linewidth, trim=50 100 100 0, clip]{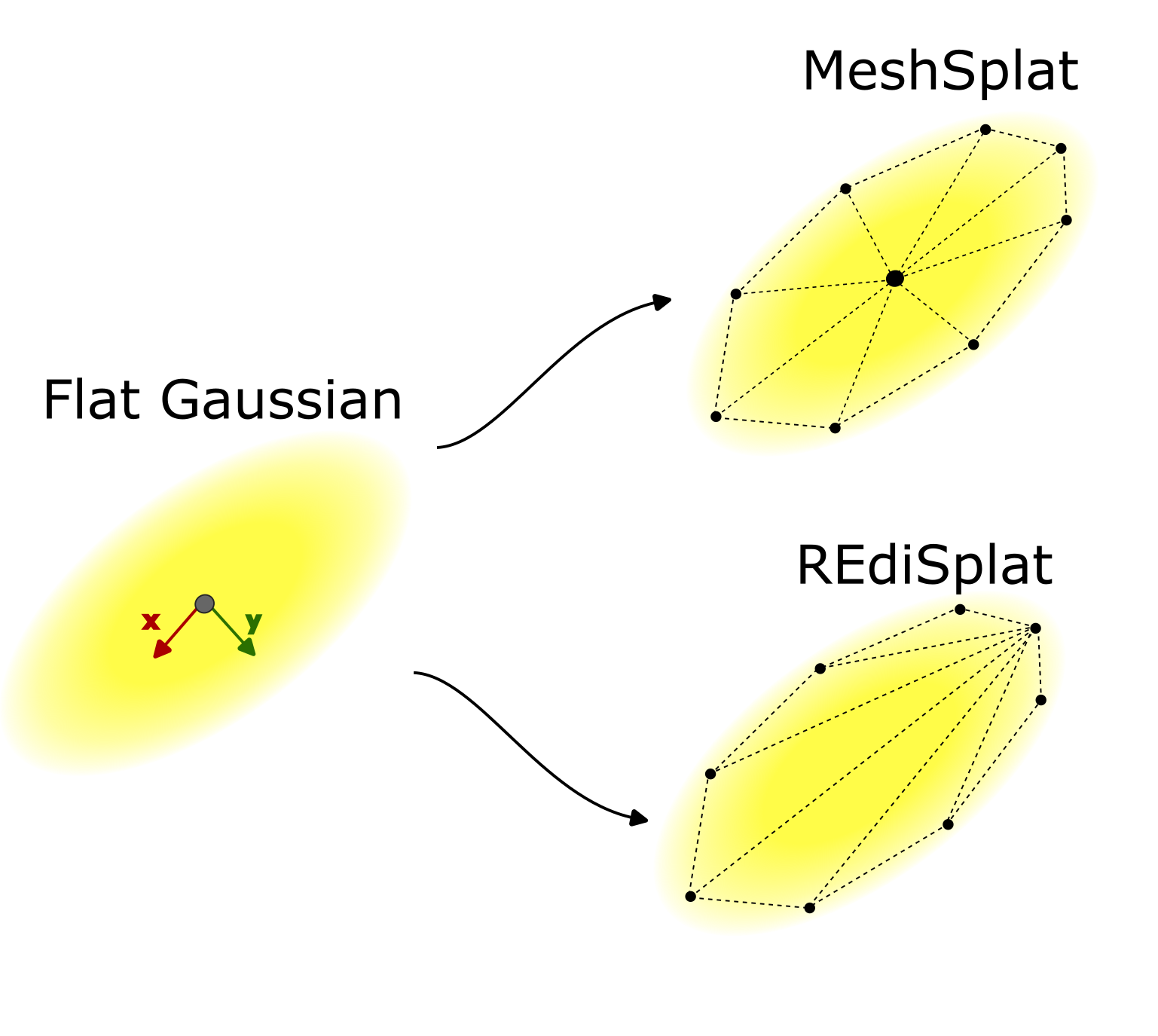}
  \end{center}  
  \vspace{-0.8cm}
    \caption{Comparison between \our{} and the MeshSplats approximation of flat Gaussian.}
    \label{fig:mesh_com}
\vspace{-0.3cm}
\end{wrapfigure}

In GS, we can use flat Guassians:
$
(\N(\m,R,S), \sigma, c ),
$
where $S=\mathrm{diag}(s_1,s_2,s_3)$, with $s_1=\varepsilon$, and $R$ is the rotation matrix defined as
$R=[\r_1,\r_2,\r_3]$, with $\r_i \in \R^3$. A mesh face can approximate a flat Gaussian as in MeshSplats \cite{tobiasz2025meshsplats}. Such an approximation can be used for ray tracing and Gaussian editing. In \our{}, we slightly modify the mesh approximation to obtain fewer faces, as shown~in~Fig.~\ref{fig:mesh_com}. 

\subsection{Mesh approximation of flat Gaussian}
\label{mesh_approximation}

We approximate each Gaussian by the planar polygon whose vertices lie on the surface spanned by the Gaussian local coordinate frame's $OY$ and $OZ$ axis. Throughout the intensive testing, we find the number of sides $n=8$ as the optimal value that constitutes the trade-off between the efficiency of the rendering and the quality of the Gaussian representation. In the above-mentioned Gaussian local coordinate frame, each vertex is given by the following formula:
$$
x_i = 0, 
\qquad
y_i = \sqrt{Q} \cos \left( \frac{2\pi}{8}i \right),
\qquad
z_i = \sqrt{Q} \sin \left( \frac{2\pi}{8}i \right),
$$
where $i=\left \lbrace 0, \ldots, 7 \right \rbrace$ and $Q = F_{\chi^2(3)}^{-1} \left( \alpha \right)$ is the quantile of order $\alpha$ of the $\chi^2(3)$ distribution (i.e., the Chi-squared distribution with three degrees of freedom) for the configurable value of $\alpha \in [0, 1)$. Since for some fixed Gaussian component: $\Sigma = R S S R^T$, in order to obtain the coordinates of the vertices $P_i \in \mathbb{R}^3$ in the global coordinate system, we have to apply to the vertices above the following affine transformation:
\begin{equation}
P_i = R{S}
\begin{bmatrix}
0 \\
\sqrt{Q} \cos \left( \frac{2\pi}{8}i \right) \\
\sqrt{Q} \sin \left( \frac{2\pi}{8}i \right)
\end{bmatrix}
+ \m
\label{points}
\end{equation}
For memory efficiency reasons, we store the vertices in the vertex buffer and utilize the triangulation of the polygon in the form of the triangle fan, which allows us to reuse each of the vertices in the vertex buffer. Below, we provide the triangle vertices triplets of the octagon constituting our approximation of the flat Gaussian:
$
\left( 0, 1, 2 \right), \left( 0, 2, 3 \right), \ldots, \left( 0, 6, 7 \right).
$

\begin{wrapfigure}{r}{0.5\textwidth}
\vspace{-0.5cm}
  \begin{center}
    \includegraphics[width=0.9\linewidth, trim=0 0 0 0, clip]{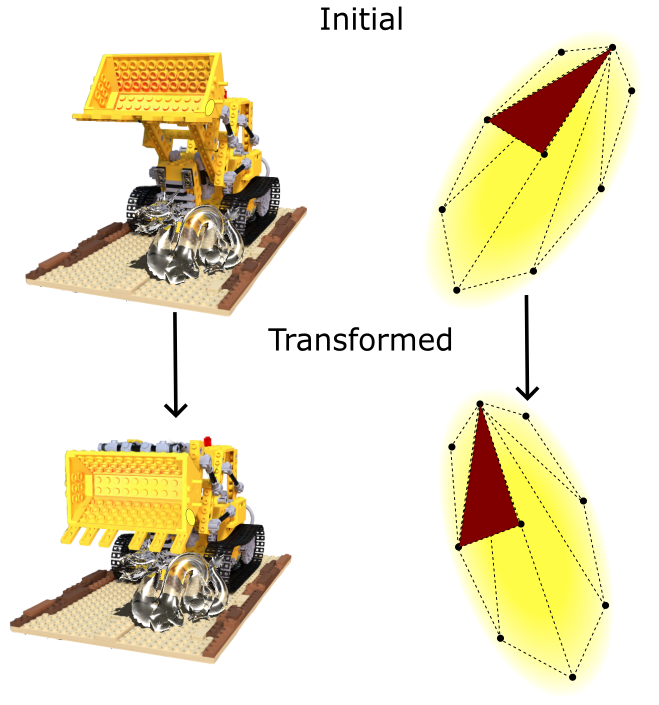}
    \caption{Our model allows transformation by editing the mesh. Unfortunately, editing all vertices produces artifacts. Therefore, we select three points (the brown triangle in the figure) that are used for the modification of each mesh. }
    \label{fig:mesh_editing}
  \end{center}
\vspace{-1.0cm}
\end{wrapfigure}

\subsection{Mesh modification}
The proposed mesh approximation can be used to edit the underlying Gaussians by moving the vertices \(P_i\) to \(P_i'\), see Fig. \ref{fig:mesh_editing}. However, the problem that might appear is the loss of imposed structure for the polygon. To alleviate this problem, we choose a set \(I=\{i_1, i_2\}\subset \{0, \ldots,n-1\}\), such that \(\#I=2\) and \(m=\frac{P_{n/4} - P_{3n/4}}{2}\) before modification, from which the parameters of a Gaussian can be restored.
In practice, we modify these three points and reconstruct a flat Gaussian.
This operation takes modified points $\{P_i'\}_{i=0}^{n-1}$ and returns \(R = \left[r_1, r_2, r_3 \right]\) and \(S=\text{diag}(s_1,s_2,s_3)\).

Such operation is given by solving the equation (\ref{points}) for \(R, S\). Let \(\tilde{P_i}=P_i'-m\) and then:
\begin{equation}
\frac{\tilde{P_i}}{\sqrt{Q}} = R S 
\begin{bmatrix}
0 \\
 \cos \left( \frac{2\pi}{8}i \right) \\
 \sin \left( \frac{2\pi}{8}i \right)
\end{bmatrix}.
\end{equation}
expanding the equality:
\begin{equation}
    \frac{\tilde{P_i}}{\sqrt{Q}} = s_2 \cos\left( \frac{2\pi}{8}i \right) r_2 + s_3 \sin\left( \frac{2\pi}{8}i \right) r_3,
\end{equation}
and taking a scalar product with regards to \(r_2\), we obtain:
\begin{align*}
    \left<\frac{\tilde{P_i}}{\sqrt{Q}}, r_2\right> &= \left< s_2 \cos\left( \frac{2\pi}{8}i \right) r_2, r_2\right> + \left< s_3 \sin\left( \frac{2\pi}{8}i \right) r_3, r_2\right> \\
    &= s_2 \cos\left( \frac{2\pi}{8}i \right) \lVert r_2 \rVert^2 = s_2 \cos\left( \frac{2\pi}{8}i \right).
\end{align*}

We define \(r_2=\frac{\tilde{P_{i_1}}} {\lVert\tilde{P_{i_1}}\rVert}\) and substituting \(r_2\) get:
\[
    s_2 = \frac{\left<\tilde{P_{i_1}}, \tilde{P_{i_1}}\right>}{\sqrt{Q}\lVert\tilde{P_{i_1}}\rVert\cos\left(\frac{2\pi}{8}i_1\right)} = \frac{\lVert\tilde{P_{i_1}}\rVert}{\sqrt{Q}\cos\left(\frac{2\pi}{8}i_1\right)}.
\]
Than, we obtain \(r_3\) by applying one step of the Gram-Schmidt process \cite{bjorck1994numerics} we get:
\[
\mathrm{orth}(\x, r_2) = x - \mathrm{proj}(\x, r_2), 
\text{where}
\;\mathrm{proj}({\bf v},{\bf u}) = \frac{ \langle {\bf v} , {\bf u} \rangle }{ \langle {\bf u},  {\bf u} \rangle } {\bf u}.
\]
\[
r_3 = \frac{\mathrm{orth}(P_{i_2}' - \m, r_2)}{ \lVert \mathrm{orth}(P_{i_2}' - \m, r_2)\rVert} \text{ and } s_3 = \frac{\left<P_{i_2}' - \m, r_3\right>}{\sqrt{Q}}
\]
Finally \(r_1 = r_2\times r_3\) and \(s_1=\varepsilon\). This preserves the parameters of each Gaussian if no transformation is done. For \(n=8\) we set \(I=\{0, 2\}\).

\subsection{Intersection Ray and Gaussian Distribution}

Our approach resembles the techniques outlined in DSS~\cite{yifan2019differentiable}, which introduces a high-fidelity differentiable renderer specifically for point clouds. The anticipated color $C({\bf r})$ of the camera ray, defined by the equation: 
$$
{\bf r}(t) = {\bf o} + t {\bf d},
$$
where ${\bf o}$ is the origin and ${\bf d}$ is the direction, is calculated by combining the colors and opacities of the Gaussians encountered by the ray. Therefore, we need the formulas for the intersection between the Gaussians and the ray. 

We can treat Gaussian as a geometric primitive. We approximate Gaussians by the confidence ellipsoid, defined by the covariance matrix $\Sigma$ and the mean vector $\m$, at a confidence level $\alpha \in [0, 1)$. Geometrically, it is the set of points $\E_{{\bf \mu}, \Sigma, \alpha}$ given by the following formula:
\begin{equation}
    \E_{{\bf \mu}, \Sigma, \alpha} = \{ {\bf x} \in \R^3 \colon ({\bf x}-{\bf \mu})^T\Sigma^{-1}({\bf x}-{\bf \mu}) = Q\},
\end{equation}
where $Q = F_{\chi^2(3)}^{-1} \left( \alpha \right)$ is the quintile of order $\alpha$ of the distribution $\chi^2(3)$ (that is, the Chi-square distribution with three degrees of freedom).

In our work, we use flat Gaussians. Therefore, the eigenvalue corresponding to the eigenvector forming the first axis of the local coordinate frame is equal to $\varepsilon~\approx~0$. In practice, our ellipses are flat, allowing us to approximate each Gaussian using a mesh of a triangulated octagon that resembles the form of these ellipses. To find the intersection between the ray and the Gaussians, we implicitly use the highly efficient built-in NVIDIA GPU RT cores primitives computing the ray-triangle intersection by leveraging the OptiX geometry triangle build type with default intersection shader to store the triangulated polygon mesh. After determining where the ray intersects the scene at its nearest point, we consider the ray to intersect with the triangle at this point. This triangle is part of an octagon that approximates a specific Gaussian. We proceed with successive steps as detailed in \cite{byrski2025raysplats}, treating this interaction as the ray that intersects the Gaussian.

\begin{figure}[!t]
    \centering
    \begin{center}
    \includegraphics[width=0.99\linewidth, trim=0 0 0 0, clip]{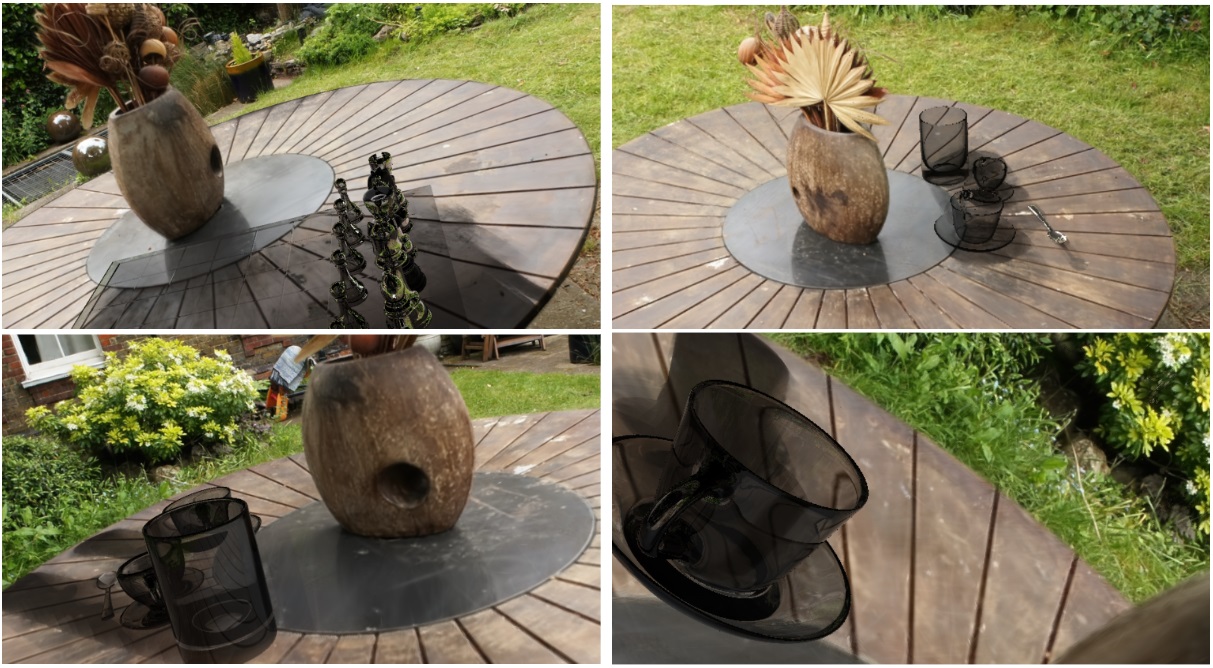}
    \end{center} 
        
    \caption{\our{} allows to model glass objects in Gaussian Splatting scene. }
    \label{fig:gs_glass}
\end{figure}

\subsection{Color Aggregation Along the Ray}

Identifying the point at which a ray intersects with a mesh that signifies a Gaussian component enables us to apply the color aggregation method as described by \cite{byrski2025raysplats,moenne20243d}.

Consider a family of Gaussians collected along the ray ${\bf r}(t) = {\bf o} + t {\bf d}$, where $\bf o$ is the origin and $\bf d$ is the direction:
\begin{equation}
    \G_{{\bf r}(t)} = \{(\N(\m_i,\Sigma_i), \hat{\alpha}_i, c_i) \}_{i=1}^{N}.
\end{equation}
Our methodology resembles 3DGRT \cite{moenne20243d} by calculating $\alpha_i$ as the product of the learned Gaussian opacity, $\hat{\alpha}_i$, and the normalized peak of the 3D Gaussian probability density function along the ray as shown in the formula:
\begin{equation}\label{equ:ray}
\alpha_i = \hat{\alpha}_i \max\limits_{t \ge 0} \left\lbrace (2\pi)^{\frac{3}{2}} \sqrt{\left \lvert \Sigma_i \right \rvert} f_{\mathcal{N} \left( \m_i, \Sigma_i \right) } \left( {\bf r}(t) \right) \right \rbrace.
\end{equation}

Since in our model we use $s_1=\varepsilon$ and approximate the very Gaussian by the triangulated mesh of the octagon, we can utilize the following practical numerically efficient approximation of the $\alpha_i$:
\begin{equation}
\alpha_i \approx \hat{\alpha}_i \cdot (2\pi)^{\frac{3}{2}} \sqrt{\left \lvert \Sigma_i \right \rvert} f_{\mathcal{N} \left( \m_i, \Sigma_i \right) } \left( {\bf r} \left(\hat{t} \right) \right),
\end{equation}
where $\hat{t}$ is the parameter value such that ${\bf r}\left( \hat{t} \right) = {\bf o} + \hat{t} {\bf d}$ is the point of intersection between the ray and the triangular mesh. Consequently:
\begin{equation}
\alpha_i \approx \hat{\alpha}_i \cdot e^{-\frac{1}{2} { \left( {\bf r} \left( \hat{t} \right) -{\bf \m_i} \right)}^T \Sigma_i^{-1} \left( {\bf r} \left( \hat{t} \right)-{\bf \m_i} \right) }
\end{equation}
Let us define: ${\bf o}' \coloneq \Sigma_i^{-1} \left( {\bf o} - m_i \right) $ and ${\bf d}' \coloneqq \Sigma_i^{-1} {\bf d}$. Then, the equation above can be further simplified as:
\begin{equation}
\alpha_i \approx \hat{\alpha}_i \cdot e^{-\frac{1}{2} { \left( {\bf o}' + \hat{t} {\bf d}' \right)}^T \left( {\bf o}' + \hat{t} {\bf d}' \right) }
\end{equation}
to finally obtain:
\begin{equation}
\alpha_i \approx \hat{\alpha}_i \cdot e^{-\frac{1}{2} { \left \lVert {\bf o}' + \hat{t} {\bf d}' \right \rVert}^2 }.
\end{equation}

For the color aggregation method, we employ a slightly modified version of the approach described in Byrski et al. \cite{byrski2025raysplats} as detailed in Algorithm~\ref{alg:coloragg}.

\begin{algorithm}[!ht]\label{alg:coloragg}
\caption{Color Aggregation in the Forward Phase}
\begin{algorithmic}[1]
\ForEach{$\left( i, j \right) \in \left \lbrace 1, \ldots, h \right \rbrace \times \left \lbrace 1, \ldots, w\right \rbrace$}
\State { \multiline{%
Compute ray $\bf o$ and $\bf d$ for the pixel index $\left( i, j \right)$ \\
in the output image;
}}
\State { $\left(T_1, T_2, I \right) \gets \left( 1, 1, 0 \right)$; }
\State { $\text{second\_phase} \gets \textbf{false}$; }
\For{$k \in \left \lbrace 1, \ldots, \text{max\_Gaussians\_per\_ray} \right \rbrace $} \label{lst:line:max_gaussians_per_ray}
\State { $\text{result} \gets \text{trace\_ray}\left( \bf o , \bf d \right) $;} \label{lst:line:ray_traversal}
\If{result \!.\! is\_empty()} \label{lst:line:intersection_is_empty}
\State { $ \text{indices}[k] \gets -1 $; } \label{lst:line:decrease_indices_buffer}
\State { \textbf{break}; } \label{lst:line:terminate_algorithm}
\Else
\State { $ \hat{t} \gets \text{result \!. \!t\_hit}$; }
\State { $ \text{index} \gets \text{result \!. \!index} / $6$ $; } \label{lst:line:save_result_index}
\State { $ \text{indices}[k] \gets \text{index} $; }
\State { $ c \gets \mathcal{G}_{\text{index},c}$; }
\State { $ \hat{\alpha} \gets \mathcal{G}_{\text{index},\hat{\alpha}}$; }
\State { Compute $\Sigma^{-1}$ for Gaussian $\mathcal{G}_{\text{index}}$; }
\State { $ \left( {\bf o}', {\bf d}' \right) \gets \left( \Sigma^{-1} \left( \bf o - \m_{\text{index}} \right), \Sigma^{-1} \bf d \right)$; }
\State { $\alpha \gets \hat{\alpha} \cdot e^{-\frac{1}{2} { \left \lVert {\bf o}' + \hat{t} {\bf d}' \right \rVert}^2 } $};
\State { $I \gets I + c \cdot \alpha \cdot T_1$; }
\State { $T_1 \gets T_1 \left( 1 - \alpha \right)$; }   \label{lst:line:update_t_1}
\If {$T_1 < \varepsilon_1$} \label{lst:line:below_threshold}
\If {$\textbf{not} \; \text{second\_phase}$}
\State { $ \text{second\_phase} \gets \textbf{true}$; } \label{lst:line:second_phase}
\Else
\State { $T_2 \gets T_2 \left( 1 - \alpha \right)$; } \label{lst:line:update_transimttance_t2}
\EndIf
\EndIf
\If {$T_2 < \varepsilon_2$} \label{lst:line:second_threshold}
\If {$k < \text{max\_Gaussians\_per\_ray}$} \label{lst:line:k_less_than_max_gaussian_per_ray}
\State { $ \text{indices}[k+1] \gets -1 $; } \label{lst:line:sentinel_value_in_indices_buffer}
\EndIf
\State { \textbf{break}; }
\Else
\State { $\bf o \gets \bf o + \varepsilon \bf d$; } \label{lst:line:move_ray_origin}
\EndIf
\EndIf
\EndFor
\EndFor
\end{algorithmic}
\end{algorithm}

While the current Gaussian count ($k$) remains below the configurable maximum (line ~\ref{lst:line:max_gaussians_per_ray}), we traverse the ray (line ~\ref{lst:line:ray_traversal}). If an intersection occurs (line ~\ref{lst:line:intersection_is_empty}), we determine the Gaussian index based on the hit triangle index belonging to the mesh constituting a triangulation of the octagon approximating the Gaussian (note that the triangulation of the octagon consists of exactly $6$ triangles) utilizing the contiguous memory layout of the consecutive triangles and we aggregate the output image color $I$ (lines ~\ref{lst:line:save_result_index}--~\ref{lst:line:update_t_1}). Otherwise, we mark the indices buffer with $-1$ for backward gradient computation (line ~\ref{lst:line:decrease_indices_buffer}) and exit (line ~\ref{lst:line:terminate_algorithm}). If, following color aggregation, transmittance $T_1$ drops below threshold $\varepsilon_1$ (line ~\ref{lst:line:below_threshold}), we activate the second phase flag (line ~\ref{lst:line:second_phase}). This triggers the second phase, where Gaussians are collected for the final significant gradient calculation or updates transmittance $T_2$ (line ~\ref{lst:line:update_transimttance_t2}) if the flag has not been set yet. Finally, we assess whether $T_2$ falls below threshold $\varepsilon_2$ (line ~\ref{lst:line:second_threshold}). If so, we terminate after marking the indices buffer (lines ~\ref{lst:line:k_less_than_max_gaussian_per_ray}--~\ref{lst:line:sentinel_value_in_indices_buffer}), or slightly adjust the ray origin o to avoid reintersecting the Gaussian at the current index in the next iteration (line ~\ref{lst:line:move_ray_origin}).


\begin{figure}[!t]
    \centering
        \begin{center}
            \our{} allows manual modifications\\
        \includegraphics[width=0.99\linewidth, trim=0 0 0 0, clip]{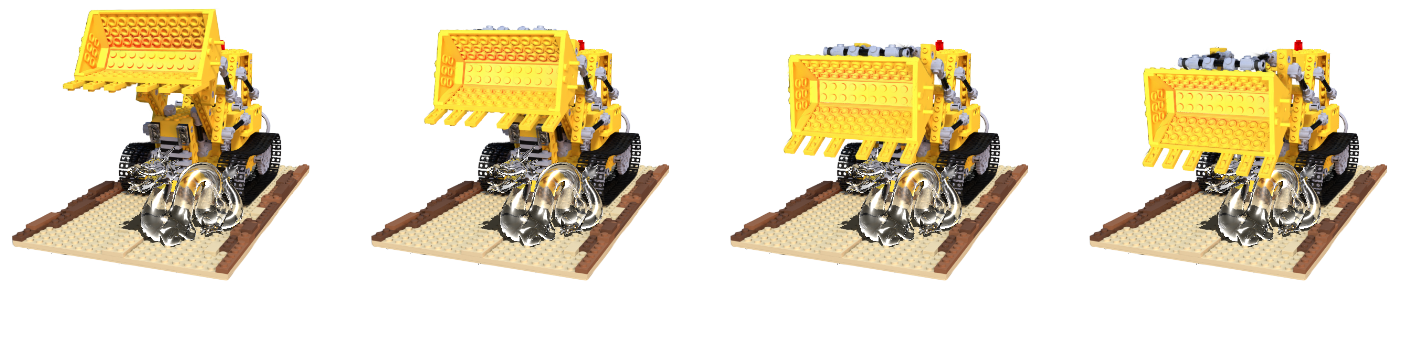}    
        \end{center} 
        \vspace{-0.7cm}
    \caption{\our{} allows manual modifications of the 3D scene. }
    \label{fig:gs_simulations}
\end{figure}

\section{Experiments}

Here, we present a comparison of our model with baseline solutions. Since our model can be used in many different scenarios, we divide this section into the following parts. We start with a classical reconstruction task, where we compare our model with previous ray tracing-based models and algorithms that work directly on meshes. Then, we demonstrate how our model cooperates with light reflection and mesh-based objects. Next, we illustrate that our model can be rendered in Blender and Nvdiffrast. Finally, we present manual modifications of our model, as well as interaction with the physics engine.

\begin{table}[]
       \caption{Quantitative evaluation of \our{} on the Mip-NeRF360 \cite{barron2022mip}, Tanks and Temples \cite{knapitsch2017tanks}, and Deep Blending \cite{hedman2018deep} datasets. We provide a comparison to the following state-of-the-art baselines: Plenoxels \cite{fridovich2022plenoxels},
    INGP \cite{muller2022instant}, M-NeRF360 \cite{barron2021mip}, 3DGS \cite{kerbl20233d}, 3DGRT \cite{moenne20243d}, LinPrim~\cite{von2025linprim}, RadiantFoam~\cite{govindarajan2025radiant}, MeshSplats\cite{tobiasz2025meshsplats}, and RaySplats~\cite{byrski2025raysplats}. On Mip-NeRF360 and Tanks and Temples, \our{} achieves comparable results to rasterization-based techniques despite its mesh-based representation. In addition, \our{} shows superior performance on Deep Blending, proving its effectiveness for indoor geometries. Note that LinPrim lacks results on the Tanks and Temples and Deep Blending datasets \cite{von2025linprim}, while RadiantFoam only lacks results on the latter \cite{govindarajan2025radiant}. 
    }

\begin{center}
{
\scriptsize
    \begin{tabular}{@{}c@{}cc@{}c@{}c@{}c@{}c@{}c@{}c@{}c@{}c@{}}
    &
    &  \multicolumn{3}{c}{Mip-NeRF360} & \multicolumn{3}{c}{Tanks and Temples} & \multicolumn{3}{c}{Deep Blending} \\
    \toprule
         & & SSIM $\uparrow$& PSNR $\uparrow$& LPIPS $\downarrow$& SSIM $\uparrow$& PSNR $\uparrow$& LPIPS $\downarrow$ & SSIM $\uparrow$& PSNR $\uparrow$& LPIPS $\downarrow$
 \\ \midrule
\multirow{8}{*}{\rotatebox{90}{ 
 \tiny Spherical Harmonics}}
&
Plenoxels &  0.670 & 23.63 &  0.44 & 0.379& 
21.08& 0.795 & 
0.510& 23.06& 0.510
\\
&
INGP-Base & 0.725 & 26.43 & - & 0.723 & 21.72 & 0.330 & 0.797 & 23.62 & 0.423
\\
&
INGP-Big & 0.751 & 26.75 &  0.30 & 0.745 & 21.92 & 0.305 & 0.817 & 24.96 & 0.390
\\
&
M-NeRF360 &   0.844 & \redc 29.23 &  - & 0.759 & \yellowc 22.22 & 0.257 & \yellowc 0.901 &  29.40 & \orangec  0.245
\\
&
3DGS-30K & \redc  0.87 & \yellowc 28.69 & \orangec  0.22  & \redc 0.841 & \orangec  23.14 &  \redc 0.183 & \orangec 0.903 & 29.41 & \redc 0.243\\
&
3DGRT & \orangec 0.854 & \orangec 28.71 & 0.25 &  \orangec 0.830 & \redc 23.20 &  0.222 & 0.900 &  29.23 &  0.315\\
&
LinPrim &  0.803 & 26.63 & \yellowc 0.221  & - & - &  - & - &  - &  - \\
&
RadiantFoam & 0.83 & 28.47 & \redc 0.21  & - & - &  - & 0.89 &  28.95 &  0.26 \\
\midrule
\multirow{3}{*}{\rotatebox{90}{ 
\tiny
RGB}}
&
RaySplats & 0.846 &  27.31 &  0.237  &  \yellowc 0.829 &  22.20 & \orangec 0.202 &  0.900 & \orangec 29.57 &  0.320\\
&
MeshSplats &  0.817 & 28.08 & 0.229  &  0.766 & 21.71 &  0.248 & 0.890 &  \yellowc 29.50 & \yellowc 0.254\\
&
\our{} (our) &  \yellowc 0.848 & 27.60 & 0.234  &  0.822 & 22.17 &  \yellowc 0.21 & \redc 0.911 &  \redc 30.01 & 0.316 \\
\bottomrule     
    \end{tabular}
    }
\end{center}    
    \label{tab:scene_full}
\end{table}

\begin{figure}[t!] 
    \makebox[\columnwidth][c]{%
        \begin{tabular*}{\dimexpr\columnwidth-1.7cm}{@{\extracolsep{\fill}} c c c c }
            \ \ \ \ GT & \ \ \ \ \ MeshSplats & \ \ \our{} (our) & 3DGS
        \end{tabular*}
    }\\[1mm] 
    \centering
    \includegraphics[width=\columnwidth]{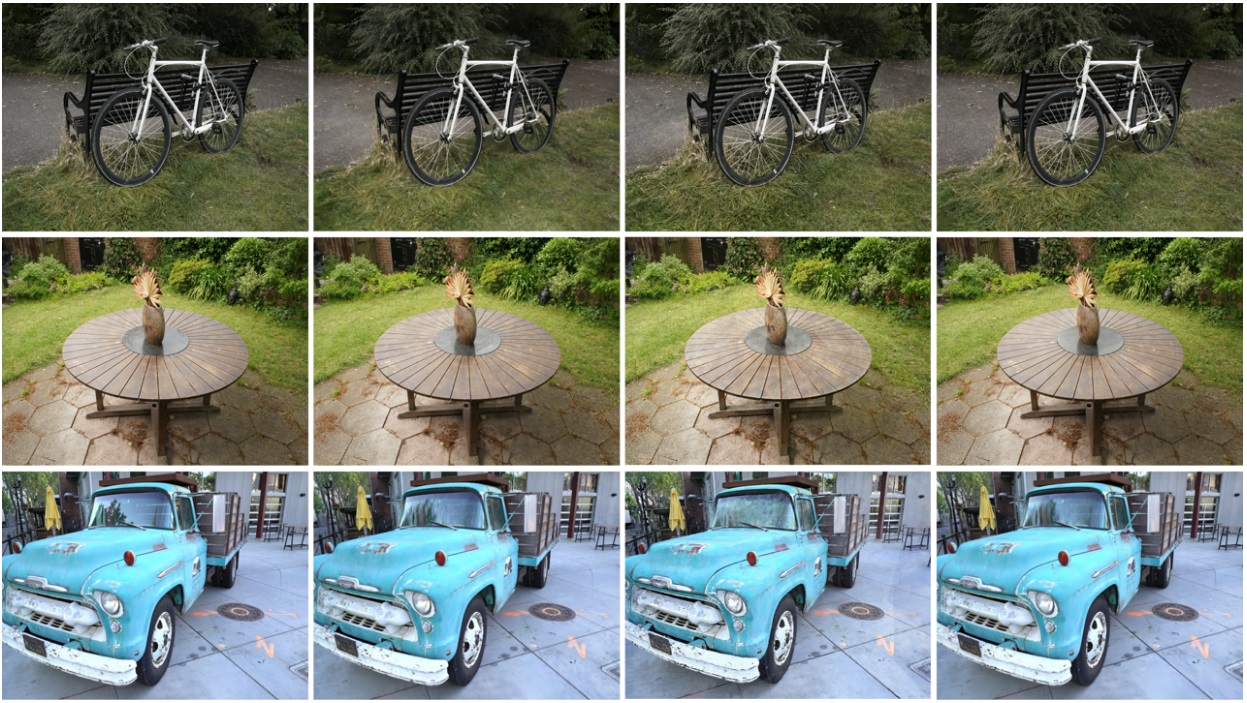}
    \caption{Our meshes can be rendered in Nvdiffrast. The first column shows the ground truth image, the second column shows the rendering of the optimized \our{} using Nvdiffrast, the third column shows MeshSplats, and the fourth column shows the rendering of 3D GS with spherical harmonics set to zero. Our model obtains quality comparable to the referenced methods. 
}
    \label{fig:big-scene_NvDiffrast}
\end{figure}

\subsection{Datasets and Metrics}
The \our{} framework was tested on three well-known datasets: MipNeRF360~\cite{barron2022mip}, Tanks and Temples~\cite{knapitsch2017tanks}, and Deep Blending~\cite{hedman2018deep}. We evaluated the same scenes as referenced in \cite{moenne20243d} to maintain consistency. In the case of Mip-NeRF360, we looked at four indoor environments (room, counter, kitchen, and bonsai) and three outdoor areas (bicycle, garden, and stump). We concentrated on two extensive outdoor scenes for the Tanks and Temples dataset: train and truck. Furthermore, the Deep Blending dataset included two indoor settings (playroom and drjohnson). In line with prior studies, we downsampled evaluation images by a factor of two for indoor scenes and four for outdoor ones. We also used consistent train/test splits across all datasets. The evaluation uses three recognized metrics: PSNR, SSIM~\cite{wang2004image}, and LPIPS~\cite{zhang2018unreasonable}.

\subsection{Quantitative Results}
For every experiment, \our{} was initialized using scenes trained for 100 iterations with GaMeS\cite{waczynska2024games}, where spherical harmonics coefficients were set to zero. This initialization ensured a stable baseline for subsequent optimization. 

As shown in Tab.~\ref{tab:scene_full}, \our{} superior results on Deep Blending dataset, outperforming state-of-the-art approaches in terms of structural similarity. This proves its effectiveness in handling complex indoor geometries and intricate scene details, where competing methods exhibit comparatively lower reconstruction quality.

For MipNeRF360 and Tanks and Temples datasets, the method delivers competitive performance, closely matching the leading baseline in reconstruction quality. Although not surpassing the highest benchmarks, its results align closely with top-performing techniques, indicating strong adaptability to both indoor and outdoor (also large-scale) environments.

Across all datasets, perceptual consistency remains comparable to established methods, reflecting balanced integration of ray tracing and mesh-based editing without compromising visual coherence. The results collectively emphasive \our{}' ability to harmonize advanced rendering features, such as light interactions, with reliable scene reconstruction.

\subsection{Qualitative Results}

\our model can work threefold: (i) as a dedicated GS renderer, with (ii) Blender, and with (iii) Nvdifrast. Quantitative comparison presented in Table~\ref{tab:scene_full} is produced in dedicated renderings using Gaussian components and ray tracing. Recently, MeshSpalts \cite{tobiasz2025meshsplats} have shown that a flat Gaussian can be represented by a mesh and rendered in Blender and Nvdifrast. Following this direction, we present visualization using these two approaches, see Fig.~\ref{fig:big-scene_NvDiffrast} and Fig.~\ref{fig:blender}.

In Fig.~\ref{fig:gs_glass}, we present how \our{} can model light reflection in glass objects. As we can see, the glass is transparent and reflects GS-based objects. Moreover, \our{} can be combined with meshes as shown in Fig.~\ref{fig:gs_mesh}. In such a case, the model shadows and light reflection. Finally, we can add mirrors (see Fig.~\ref{fig:gs_mirror}), which perfectly reflect the entire GS-based scene. Moreover, 3D scenes modeled by \our{} can be edited, as presented in Fig.~\ref{fig:gs_simulations}.

We generate visualizations using Blender and Nvdiffrast. In the case of the former, physical simulations are depicted in Fig~\ref{fig:blender}. We also compare our model with MeshSplats in Nvdiffrast, as shown in Fig.~\ref{fig:big-scene_NvDiffrast}. Renders obtained by \our{} have similar quality to MeshSplats.
Our model provides results similar to GS and can be represented by a mesh akin to MeshSplats.

\section{Conclusions}

GS enables high-quality neural rendering but struggles with lighting variations, physical interactions, and manual adjustments. To address these issues, we introduce \our{}, which combines ray tracing with a mesh-based representation of flat 3D Gaussians. Thus allowing better control over lighting, modifications, and physical simulation. Additionally, our approach integrates with standard 3D tools like Blender and Nvdiffrast, enhancing compatibility with traditional graphics pipelines. \our{} bridges the gap between GS and conventional 3D rendering, offering a more flexible and physically accurate framework for future advancements.

\begin{credits}
\subsubsection{\discintname}
The authors have no competing interests to declare that are relevant to the content of this article. 
\end{credits}

%
%


%
\bibliographystyle{splncs04}

%





\end{document}